\documentclass[letterpaper]{article} 
\usepackage{aaai25}  
\usepackage{times}  
\usepackage{helvet}  
\usepackage{courier}  
\usepackage[hyphens]{url}  
\usepackage{graphicx} 
\usepackage{natbib}  
\usepackage{caption} 
\usepackage[table]{xcolor}

\usepackage{subcaption}
\usepackage{booktabs}
\usepackage{multirow}
\usepackage{amsmath}
\usepackage{amssymb}
\usepackage{array}
\usepackage{pifont}
\usepackage{enumitem}
\usepackage{algorithm}
\usepackage{algorithmic}
\definecolor{mygray}{RGB}{220,220,220}
\newcommand{\cmark}{\ding{51}} 
\newcommand{\xmark}{\ding{55}} 

\usepackage{newfloat}
\usepackage{listings}
\DeclareCaptionStyle{ruled}{labelfont=normalfont,labelsep=colon,strut=off} 
\floatstyle{ruled}
\newfloat{listing}{tb}{lst}{}
\floatname{listing}{Listing}
\title{ Enhancing Instruction-Following Capability of Visual-Language Models by Reducing Image Redundancy}

\author {
    Te Yang\textsuperscript{\rm 1,2},
    Jian Jia\textsuperscript{\rm 3},
    Xiangyu Zhu\textsuperscript{\rm 1,2},
    Weisong Zhao\textsuperscript{\rm 4},
    Bo Wang\textsuperscript{\rm 3},
    Yanhua Cheng\textsuperscript{\rm 3},
    Yan Li\textsuperscript{\rm 3},
    Shengyuan Liu\textsuperscript{\rm 2},
    Quan Chen\textsuperscript{\rm 3},
    Peng Jiang\textsuperscript{\rm 3},
    Kun Gai\textsuperscript{\rm 3},
    Zhen Lei\textsuperscript{\rm 1,2,5}
}
\affiliations {
    \textsuperscript{\rm 1}State Key Laboratory of Multimodal Artificial Intelligence Systems,
Institute of Automation, Chinese Academy of Sciences \\
    \textsuperscript{\rm 2}School of Artificial Intelligence, University of Chinese Academy of Sciences, Beijing, China\\
    \textsuperscript{\rm 3}Kuaishou Technology\\
    \textsuperscript{\rm 4}School of Cyber Security, University of Chinese Academy of Sciences, Beijing, China\\
    \textsuperscript{\rm 5}CAIR, HKISI, Chinese Academy of Sciences\\
    \{yangte2021, xiangyu.zhu, zhen.lei\}@ia.ac.cn, jian.jia@outlook.com,
    zhaoweisong@iie.ac.cn, wangbo0060@163.com, chengyanhua@kuaishou.com,
    yan.li@cripac.ia.ac.cn, myctllmail@163.com, jp2006@139.com,
    kun.gai@qq.com
}

\begin{document}
\maketitle
\begin{abstract}
Large Language Models (LLMs) have strong instruction-following capability to interpret and execute tasks as directed by human commands. 
Multimodal Large Language Models (MLLMs) have inferior instruction-following ability compared to LLMs. However, there is a significant gap in the instruction-following capabilities between the MLLMs and LLMs.
In this study, we conduct a pilot experiment, which demonstrates that spatially down-sampling visual tokens significantly enhances the instruction-following capability of MLLMs. This is attributed to the substantial redundancy in visual modality. However, this intuitive method severely impairs the MLLM's multimodal understanding capability.
In this paper, we propose Visual-Modality Token Compression (VMTC) and Cross-Modality Attention Inhibition (CMAI) strategies to alleviate this gap between MLLMs and LLMs by inhibiting the influence of irrelevant visual tokens during content generation, increasing the instruction-following ability of the MLLMs while retaining their multimodal understanding capacity. 
In VMTC module, the primary tokens are retained and the redundant tokens are condensed by token clustering and merging.
In CMAI process, we aggregate text-to-image attentions by text-to-text attentions to obtain a text-to-image focus score. Attention inhibition is performed on the text-image token pairs with low scores.
Our comprehensive experiments over instruction-following capabilities and VQA-V2 \cite{goyal2017making}, GQA \cite{hudson2019gqa}, TextVQA \cite{singh2019towards} , MME \cite{fu2024mmecomprehensiveevaluationbenchmark} and MMBench \cite{liu2023mmbench} five benchmarks, demonstrate that proposed strategy significantly enhances the instruction following capability of MLLMs while preserving the ability to understand and process multimodal inputs.
\end{abstract}
\section{Introduction}
\label{sec:intro}
\begin{figure}[t]
    \centering
    \includegraphics[width=0.46\textwidth]{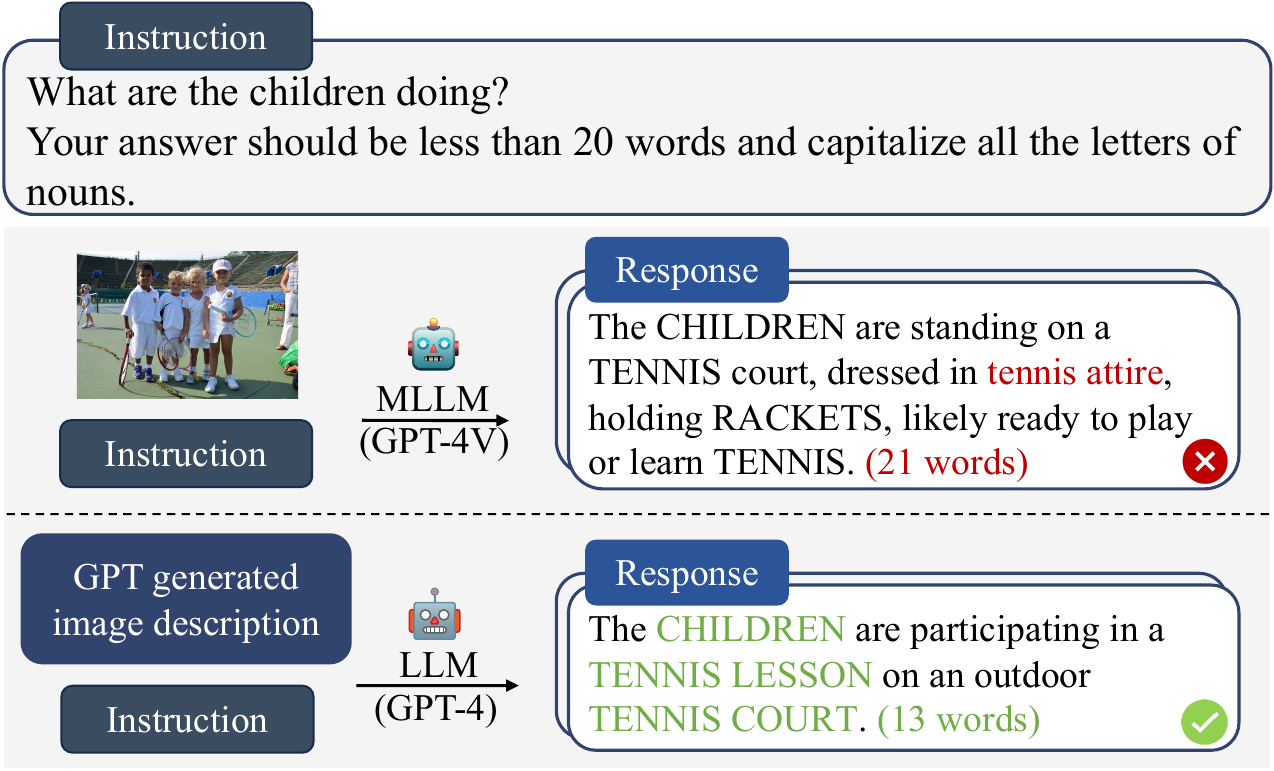}
    \vspace{-3mm}
    \caption{Illustration of the gap in the instruction-following ability between GPT4 \cite{achiam2023gpt} and GPT-4V \cite{gpt4v}. Under the same instruction, GPT-4V cannot generate outputs that meet the requirements of the instruction. 
    However, when the image in the input is replaced with a detailed description of the image generated by GPT-4, the model can correctly follow the provided instructions.  }
    \label{fig:cover}
    \vspace{-5mm}
\end{figure}

Achieving alignment between artificial intelligence systems and human intentions has long been an important objective in AI research \cite{ji2023ai, han2022aligning}. It ensures that the behavior and decision making of AI systems are consistent with human intentions and values, thereby ensuring that the development and application of AI are beneficial and pose no harm to humans. 
Ensuring that machines can precisely follow human instructions is a preliminary yet crucial step in achieving this alignment. \\
Large language models (LLMs) have been significantly improved owing to innovations in model architecture and large-scale pre-training \cite{achiam2023gpt,touvron2023llama2}, resulting in accurate responses to complex human instructions.
Despite the significant progress in MLLMs driven by advancements in LLMs, there is a substantial gap between MLLMs and their foundations LLMs regarding precise instruction-following ability.
Replacing multimodal inputs with text-only inputs significantly increases the instruction-following capabilities of MLLMs (Figure \ref{fig:cover}). This phenomenon is common and cannot be avoided even in the best closed-source MLLMs.
 \\
This gap prompts our investigation into the instruction-following capability of MLLMs.
As discussed in \cite{he2022masked}, one significant difference between the vision modality and the language modality is the extent of information redundancy.
Language is a medium of human communication with a low information redundancy and rich semantics. Conversely, images are highly spatially redundant. Notably, masked image modeling can adopt a significantly larger masking ratio, reaching 90\%, compared to language \cite{he2022masked}. 
Therefore, it is imperative to explore whether reducing the redundant information in images can enhance the instruction-following capabilities of MLLMs. \\
Conventional MLLMs generally adhere to a design paradigm comprising two stages: visual-modality processing and cross-modality content generation. In the former stage, images are tokenized using a visual encoder with an adapter. In the cross-modality content generation stage, the LLM produces the desired output using visual tokens and text embeddings as inputs.
Consequently, we formulate two research questions: 1) How can we reduce the redundancy in visual modality? 2) How can we minimize the impact of redundancy in visual modality on the cross-modality content generation process? \\
The instruction-following capabilities of MLLMs can be significant improved by reducing redundant information in images using an intuitive spatial down-sampling strategy over visual tokens.
However, this simple strategy significantly impairs visual understanding capabilities of MLLMs. 
In this study, we utilize 1) visual-modality token compression (VMTC) and 2) cross-modality attention inhibition (CMAI) to address those two above-mentioned problems, respectively.
The VMTC module is designed to compress redundant information of images while retaining the critical foreground information.
It leverages attention scores in the ViT layers to identify redundant background tokens, which are clustered based on token similarity and fused accordingly.
The CMAI module is presented to mitigate the impact of visual redundancy by ensuring that each text token in the LLM concentrates exclusively on the relevant image tokens. 
This is achieved by attenuating the attention level between text-image token pairs with low text-to-image focus scores. \\
The comprehensive experimental results demonstrate that our method achieves SOTA performance on instruction-following capabilities while precisely maintaining the multimodal understanding capabilities of MLLMs.
{The major contributions of this study are summarized as below:}
\begin{itemize}[noitemsep]
\item To the best of our knowledge, this is the first study to investigate the instruction-following capability of MLLMs from a model perspective and propose a correlation between the instruction-following capability of MLLMs and the redundancy of visual modality.
\item We propose visual-modality token compression strategy to compress redundant visual information and cross-modality attention inhibition approach to reduce the impact of visual redundancy on text generation.
\item By integrating VMTC and CMAI, our method significantly improves the instruction-following capabilities of MLLMs, while precisely maintaining the performance of the baseline model.
\end{itemize}
\vspace{-2mm}
\section{Related Work}
\textbf{Multimodal Large Language Models}.
Benefiting from recent advancements of LLMs \cite{achiam2023gpt, touvron2023llama, touvron2023llama2}, multimodal large language models have demonstrated remarkable capabilities across various visual-language tasks \cite{li2023blip, alayrac2022flamingo,dai2023instructblip,liu2024visual}. 
BLIP2 and InstructBLIP \cite{li2023blip, dai2023instructblip} employ the Q-Former to aggregate visual features and LLaVA \cite{liu2024visual} uses a linear projection layer to align the token dimensions of the visual encoder with LLM. These methods significantly enhancing the model's ability to interpret and understand images. 
Many works continue to explore this area from different directions, including adopting higher image resolutions \cite{lin2023sphinx,li2023otterhd, xu2024llava}, efficient MLLMs \cite{chu2024mobilevlm}, and extending applications to the video domain \cite{lin2023video, kim2024image}. \\
\textbf{Token Pruning.}
Researchers have proposed various approaches \cite{rao2021dynamicvit,wei2023joint, bolya2023tome}  to remove redundant tokens in ViT \cite{dosovitskiy2020image} to improve the model's computational efficiency. Meanwhile, many works \cite{cao2024madtp,cao2023pumer,wang2023smarttrim} have improved the computational efficiency of visual-language models through pruning, but these efforts have primarily focused on traditional VLMs \cite{li2022blip,radford2021learning}.
Recently, LLaVA-Prumerge \cite{shang2024llava} proposes to reduce visual tokens in last hidden layer of ViT. 
However, this work also aims to enhance the computational efficiency of MLLMs, while our focus is on improving the instruction-following capability of MLLMs.\\
\textbf{Instruction-Following.}
A series of works have been proposed to address the instruction-following capability of large language models, including human evaluation \cite{ouyang2022training,zheng2024judging, alpaca}, model-based evaluation \cite{chang2023survey,liu2023gpteval}, and evaluation based on quantitative benchmarks \cite{koubaa2023gpt,katz2024gpt, chang2023survey}. In particular, IFEval \cite{zhou2023instruction} introduces instruction-following evaluation based on verifiable instructions, which can automate the evaluation process and enhances the accuracy and consistency of the evaluation process. In the context of multimodal large models, most works \cite{li2023videochat, luo2023valley,liu2024visual}start from a data perspective, constructing instruction datasets to enable models to acquire instruction-following capability. In our study, we investigate the factors influencing the instruction-following capability of MLLMs and enhance the model's instruction-following capability from a model architecture perspective.

\section{Pilot Studies}
\begin{figure*}[ht]
    \centering
    \begin{subfigure}{0.23\textwidth}
    \includegraphics[width=\textwidth]{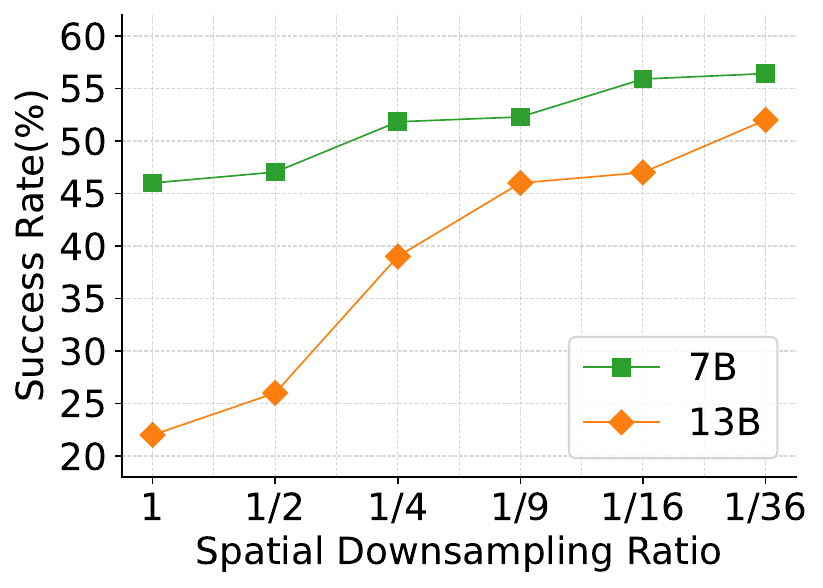}
    \caption{Instruction following performance on JSON task .}
    \label{fig:json}
    \end{subfigure}
    \begin{subfigure}{0.23\textwidth}
    \includegraphics[width=\textwidth]{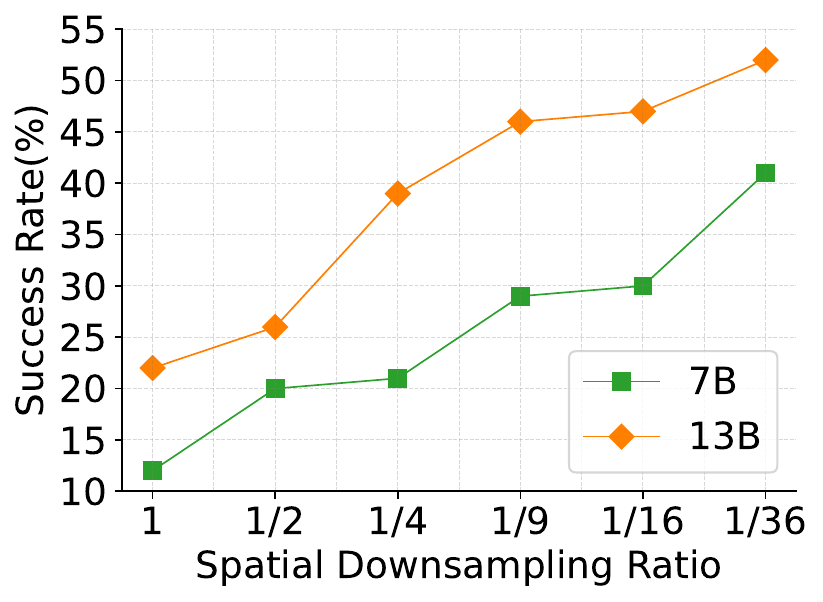}
    \caption{Instruction following performance on keywords task}
    \label{fig:keywords}
    \end{subfigure} 
    \begin{subfigure}{0.23\textwidth}
        \includegraphics[width=\textwidth]{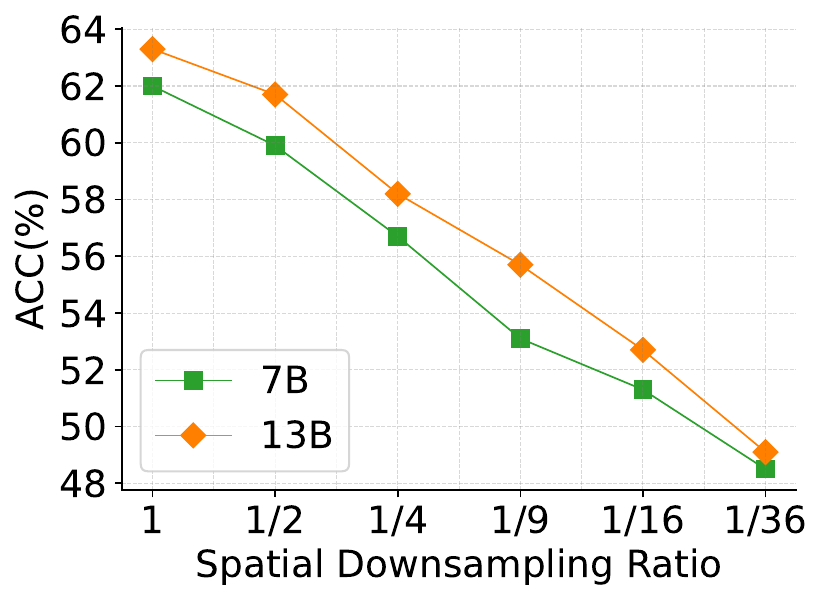}
        \caption{Multimodal understanding capability on GQA benchmark.}
        \label{fig:gqa}
    \end{subfigure}
    \begin{subfigure}{0.23\textwidth}
        \includegraphics[width=\textwidth]{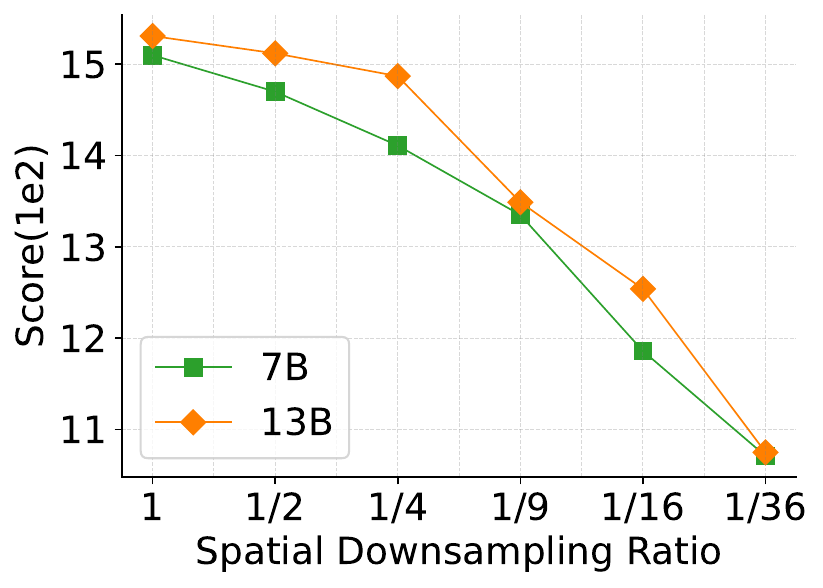}
        \caption{Multimodal understanding capability on MME benchmark.}
        \label{fig:mme}
    \end{subfigure}   
    \vspace{-3mm}
    \caption{Illustration of instruction following performance and multimodal understanding capability of LLaVA-1.5 using different spatial down-sampling ratio. In Figure \ref{fig:json} and \ref{fig:keywords}, the instruction-following performances are significantly improved as the down-sampling ratio increases. In Figure \ref{fig:gqa} and \ref{fig:mme}, down-sampling operations significantly affect the model's multimodal understanding capability.}
    \label{fig:pilot_exp}
    \vspace{-5mm}
\end{figure*}
We conduct pilot experiments on the instruction following abilities of existing MLLMs, using GPT-4V\cite{gpt4v} and LLaVA-1.5\cite{liu2024improved} as representative models.
Developed and maintained by OpenAI, GPT-4V is one of the most advanced proprietary LMM. Currently, LLaVA-1.5 is one of the most impactful open-source MLLMs available.
GPT-4V has a proprietary nature, thus we use case studies to evaluate its instruction-following capabilities. 
We manually design instructions and use them as input with a related image to assess the ability of GPT-4V model to generate accurate responses. 
GPT-4V is also used to generate descriptions of images. Subsequently, these descriptions are used as input along with the instructions to evaluate the instruction-following capabilities of GPT-4 model, as a control.
The findings demonstrated a significant gap in instruction-following capabilities between GPT-4V and GPT-4 models as shown in Figure \ref{fig:cover}. More examples are presented in the appendix.\\
Such gap prompts us to explore the differences between visual modality and language modality.
Images are unstructured information with a high degree of spatial redundancy. Conversely, text is highly structured information with low redundancy. 
As illustrated in \cite{he2022masked, wettig2022should}, when the mask ratio exceeds 80\%, the accuracy loss in masked image modeling methods is negligible (0.5\%), compared to the optimal ratio. 
In contrast, masked language modeling methods exhibit a significant degradation in accuracy (over 10\%) under similar conditions. 
This disparity indicates the differences in information redundancy between language and images. We hypothesize that reducing redundancy in images can improve the instruction-following ability of multimodal large language models. \\
\begin{figure*}[ht]
    \centering \includegraphics[width=0.9\textwidth]{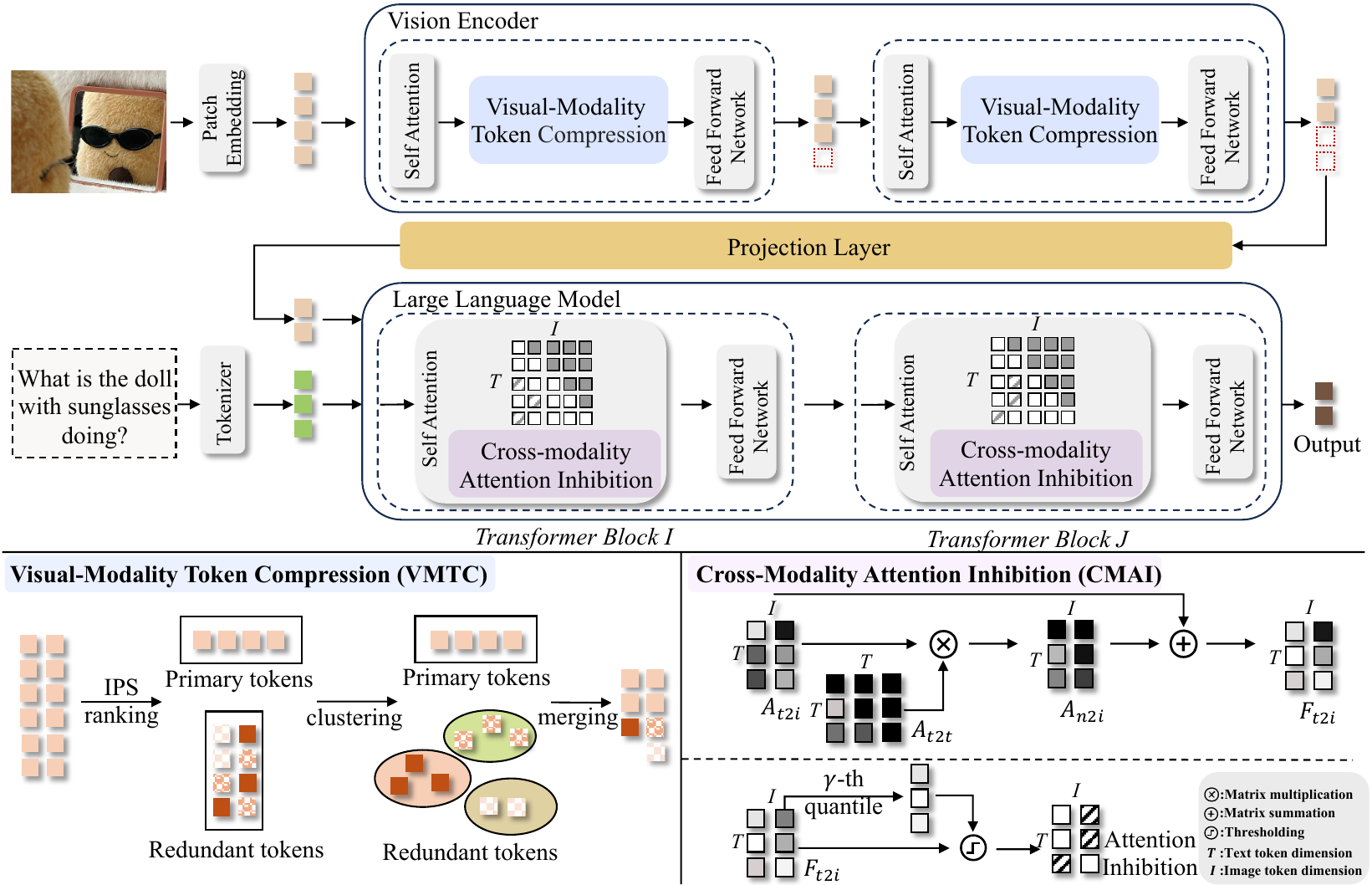}
    \vspace{-2mm}
    \caption{
    Overview of our proposed framework. Our proposed framework incorporates two principal components: the VMTC module and the CMAI module. The VMTC module is strategically positioned within several transformer blocks of the visual encoder, specifically between the self-attention and feed-forward networks. This module aims to compress image redundancy by retaining primary tokens while clustering and merging redundant tokens. Meanwhile, the CMAI module is integrated into the self-attention modules of transformer blocks in LLMs, effectively inhibiting the influence of redundant image tokens.}
    \label{fig:main_figure}
    \vspace{-5mm}
\end{figure*} 
We conduct experiments using LLaVA-1.5 \cite{liu2024improved} 7B and 13B models. 
In the pilot experiments, we conduct spatial down-sampling of the tokens obtained from the image encoded by ViT to reduce image redundancy.
We evaluate the model's instruction-following capability based on its ability to perform two simple tasks: responding in JSON format and including specific keywords in the answers.
The results of the model's instruction-following capability and multimodal understanding capability under different spatial down-sampling ratios are shown in Figure \ref{fig:pilot_exp}. 
The results indicate that an increase in the spatial down-sampling ratio reduces the redundancy in the images, leading to a significant enhancement of the model's instruction-following capabilities (Figure \ref{fig:json} and Figure \ref{fig:keywords}). 
However, the results also demonstrate that the straightforward strategy of spatial down-sampling significantly impairs the model's multimodal understanding capabilities (Figure \ref{fig:gqa} and Figure \ref{fig:mme}). 
This finding prompts us to explore more optimal solutions that enhance the instruction-following capabilities of MLLMs without substantially compromising its inherent multimodal understanding capabilities.

\section{Method}
The enhancement of instruction-following capability in MLLMs through the reduction of redundant information via spatial down-sampling in images has been previously discussed.
However, this approach significantly compromises the model's multimodal understanding ability.
Our objective is to improve the instruction-following capability of MLLMs while preserving their multimodal understanding abilities. Figure \ref{fig:main_figure} illustrates an overview of our proposed model architecture. 
This section begins with an introduction to MLLMs. The Visual-Modality Token Compression (VMTC) method is then elucidated. This technique preserves crucial foreground information while compressing less significant background details. Finally, the Cross-Modality Attention Inhibition (CMAI) is presented. CMAI is designed to mitigate the impact of redundant image information on text generation.
\subsection{Preliminary}
\label{sec:preliminary}

MLLMs typically process both image and text inputs, generating textual responses as output. 
The current predominant architecture of MLLMs generally comprises three key components: a visual encoder, a projection layer, and a LLM.
The visual encoder $E$, typically employing the Vision Transformer (ViT) \cite{dosovitskiy2020image} architecture, consists of multiple stacked transformer blocks. These blocks transform input image patches $I$ to visual tokens $T$. 
Each transformer block is composed of Multi-head Self-attention (MSA) and Feed-forward Layer components.
\begin{align}
    z'_l &= \text{MSA}(z_{l-1}) + z_{l-1}, \\
    z_l  &= \text{MLP}(z'_l) + z'_l,
\end{align}
where $z_{l-1}$ and $z_l$ are hidden states in $l-1$th and $l$th layer.
The projection layer $Proj$ is designed to align the dimensions of visual tokens with the input dimensions of the LLM.
\begin{equation}
T' = Proj(T).  
\end{equation}
As a result, the dimension of $T' \in \mathbb{R}^{n \times d}$ is aligned with the LLM.
The input of LLM consists aligned image tokens $T'$ and text $X_q$, and the output are the textual tokens $X_a$. The LLM also adopts the transformer architecture \cite{vaswani2017attention} but are typically with causal attention masks to ensure no information leakage occurs during text generation.
\begin{equation}
    \text{SelfAttn}(Q, K, V) = ( \text{softmax}(\frac{QK^T}{\sqrt{d_k}}) + M )V,
\end{equation}
where $Q,K,V$ are different projections of the hidden state $z_l$, and $M$ is the causal attention mask which has negative infinity above the diagonal and zeros in all other positions. 
 \(QK^T\) is referred to as the attention score, while the result of this expression \(\text{softmax}(\frac{QK^T}{\sqrt{d_k}})\) is the attention weights.
\subsection{Visual-Modality Token Compression}
In order to compress spatially redundant information in images and enhance the model's instruction-following capability, we implement a strategy that preserves essential foreground tokens while clusters and merges the remaining ones. This approach is based on researches \cite{wei2023joint} indicating that the complete elimination of remaining tokens can have an adverse effect on model performance.
Specifically, given the average attention weight of different attention heads $A_w \in \mathbb{R}^{(n+1)\times(n+1)}$, we define the attention weights between each patch token $\{{z'}_l^{1}, ..., {z'}_l^{n}\}$ and class token $\{{z'}_l^{0}\}$ as importance score $\text{IPS} \in \mathbb{R}^n$. Supposing the number of output tokens is $k+1+c$, where $c$ is the clustering number, the class token $\{{z'}_l^{0}\}$ and the top $k$ patch tokens with the highest importance scores $\text{IPS}$ are selected as primary tokens ${z'}_{l,kp}$ and other tokens are considered as redundant tokens ${z'}_{l,rd}$:
\begin{align}
    {z'}_{l,kp} &= \{ {z'}_l^{i} | \text{rank}(\text{IPS}(i)) \leq k, 1\leq i \leq n \}.\\
    {z'}_{l,rd} &= \{ {z'}_l^{j} | \text{rank}(\text{IPS}(j)) > k, 1\leq j \leq n \}.
\end{align}
To effectively preserve information from redundant tokens while minimizing image redundancy, a token merging strategy is employed. Given that redundant tokens may belong to diverse semantic categories, indiscriminate merging could lead to semantic confusion. To address this challenge, the K-Means algorithm is utilized to cluster tokens based on cosine similarity. Within each cluster $\{ C_1, ..., C_s\}$, tokens are considered semantically similar and are merged based on $\text{IPS}$. This approach enables the compression of redundant information while maintaining the semantic integrity of the image content. 
\begin{align}
    {z'}_{l,C_{i}}  = \sum_{ \{ j \mid z'^{j}_{l} \in z'_{l, rd}, z'^{j}_{l} \in C_{i} \} }\text{IPS}(j)  {z'}_l^{j}.
\end{align}
Finally, the intermediate representation of $z'_l$ is updated as:
\begin{equation}
    z'_l = \text{cat}({z'}_l^0, {z'}_{l,kp}, {z'}_{l,C_{1}}, ... ,{z'}_{l,C_s}).
\end{equation}

\subsection{Cross-Modality Attention Inhibition}
Despite the compression of image redundancy in the visual modality, retaining an insufficient number of tokens can significantly impair the model's multimodal understanding capability. To address this issue, we propose cross-modality attention inhibition, a plug-and-play module for Large Language Models (LLMs). This module enables each text token to focus exclusively on the most relevant image tokens, thereby mitigating the impact of redundant information. \\
Given a sequence of tokens $\{I_1, ..., I_n, T_1,...,T_m\}$, where $I$ and $T$ represent image and text tokens, respectively, and an average attention score $A_s \in \mathbb{R}^{(n+m) \times (n+m)}$, we define text-to-image attentions $A_{t2i} \in \mathbb{R}^{m \times n} \subset A_s$ as elements where the queries are text tokens and the keys are image tokens. Similarly, text-to-text attentions $A_{t2t} \in \mathbb{R}^{m \times m} \subset A_s$ are defined as elements where both the query and the key are text tokens.
The primary objective is to effectively identify and suppress incorrect associations between image and text tokens, ensuring the preservation of only relevant connections. To achieve this goal, we consider not only each text token's own attentiveness to image tokens, but also the attention of other text tokens it attends to. We use the neighborhood text-to-image attention $A_{n2i}$ to quantify this attentiveness, aggregating $A_{t2i}$ through neighborhood text-to-text attention $A_{t2n}$. 
The neighborhood text-to-text attention serves to quantify a text token's focus on neighboring tokens, excluding itself.
Finally, we combine $A_{n2i}$ and $A_{t2i}$ to obtain the text-to-image focus score $F_{t2i}$.
\begin{align}
    A_{t2n}^{jk} &= 
    \begin{cases} 
    A_{t2t}^{jk}, & \text{if } j > k, \\ 
    0, & \text{if } j \leq k,
    \end{cases} \\
    A_{n2i}^{jk} & = \sum_{h \in \{1, ..., m\}}{A_{t2n}^{jh}  A_{t2i}^{hk}} ,\\
    F_{t2i} &= A_{n2i} + A_{t2i},
\end{align}
where \( j, k \) denote the row and column indices, respectively.
Conceptually, if a text token allocates greater attention to another particular text token, then the focus of that particular text token on image tokens should be weighted more prominently in $A_{n2i}$.
Finally, given attention inhibition ratio \(\gamma\), we calculate \(\gamma\)-th quantile for each row of \(F_{\text{t2i}}\) as thresholds. Then we perform attention inhibition on text-image token pairs with \(F_{\text{t2i}}\) below these thresholds by adding negative infinities on causal masks at corresponding positions.

\section{Experiments}
\begin{table*}[t]
    \centering
    \scalebox{0.80}{
    \begin{tabular}{l|c |cccccccccccccccc|c}
    
    \toprule
    Method & GQA & T1  & T2  & T3  & T4  & T5  & T6  & T7  & T8  & T9  & T10  & T11  & T12  & T13  & T14  & T15  & T16  & AVG  \\
    \midrule
    LLaVA-1.5-7B &\underline{62.0}& 57.3 & 06.0 & 39.9 & 16.1 & 40.4 & 53.2 & 21.1 & 06.4 & 45.4 & 54.1 & 26.6 & 56.4 & 51.4 & 01.8 & 11.0 & 67.4 & 34.7  \\
    LLaVA-1.5-13B&\textbf{63.3}& 53.7 & 46.8 & 35.8 & 14.2 & 62.8 & 72.0 & 12.4 & 02.8 & 21.6 & 83.0 & 56.4 & 65.1 & 50.0 & 20.2 & 19.7 & 57.8 & 42.1  \\
    \midrule
    SPD-7B &59.9 &56.9 & 07.3 & 46.8 & 17.9 & 41.7 & 63.8 & 22.0 & 08.3 & 47.7 & 56.0 & 35.8 & 63.8 & 54.1 & 02.8 & 15.1 & 73.9 & 38.4 \\
    SPD-13B&61.7 &58.7 & 58.7 & 51.8 & 20.6 & 63.3 & 75.2 & 15.1 & 01.8 & 28.0 & 83.0 & 70.2 & 64.7 & 52.8 & 13.8 & 33.9 & 67.4 & \underline{47.4} \\
    \midrule
    \rowcolor{mygray} \color{gray} Upper Limit-7B & \color{gray}48.52 &\color{gray}35.8 &\color{gray} 01.4 &\color{gray} 38.1 &\color{gray} 33.9 &\color{gray} 68.4 &\color{gray} 80.7 &\color{gray} 33.9 &\color{gray} 10.1 &\color{gray} 68.8 &\color{gray} 90.8 &\color{gray} 62.4 &\color{gray} 66.1 &\color{gray} 53.7 &\color{gray} 07.8 &\color{gray} 42.2 &\color{gray} 100.0 &\color{gray} 49.6 \\
    \rowcolor{mygray} \color{gray} Upper Limit-13B& \color{gray}48.68&\color{gray}49.5 &\color{gray} 83.9 &\color{gray} 76.2 &\color{gray} 12.8 &\color{gray} 79.8 &\color{gray} 38.1 &\color{gray} 33.5 &\color{gray} 04.1 &\color{gray} 77.1 &\color{gray} 89.0 &\color{gray} 69.7 &\color{gray} 76.6 &\color{gray} 60.1 &\color{gray} 02.8 &\color{gray} 79.4 &\color{gray} 96.3 &\color{gray} 58.1 \\
    \midrule
    
    Ours-7B &61.6 &63.3 & 09.6 & 60.6 & 21.6 & 47.3 & 64.7 & 23.4 & 12.4 & 54.6 & 72.9 & 42.2 & 69.7 & 60.6 & 03.2 & 18.4 & 81.2 & 44.1  \\
    Ours-13B & \textbf{63.3} &63.8 & 62.8 & 69.3 & 21.0 & 65.1 & 83.0 & 13.8 & 03.7 & 28.4 & 85.3 & 74.8 & 71.6 & 56.9 & 20.6 & 37.2 & 68.8 & \textbf{51.6}  \\
    \bottomrule
    \end{tabular}}
    \vspace{-2mm}
    \caption{Comparison experimental results of Instruction Following Capabilities and GQA \cite{hudson2019gqa}. The SPD method refers to the spatial down-sampling version of LLaVA-1.5 \cite{liu2024improved}.}
    \vspace{-3mm}
    \label{tab:instruciton}
\end{table*}

\begin{table*}[t!]
\centering
\scalebox{0.80}{
\begin{tabular}{l l c c c |c| c  c  c c c}
\toprule
\multirow{2}{*}{Method} & \multirow{2}{*}{LLM} & Image & \multicolumn{2}{c|}{Sample Size} & \multirow{2}{*}{IF $\uparrow$} & \multirow{2}{*}{VQAv2$\uparrow$} & \multirow{2}{*}{GQA$\uparrow$}   & \multirow{2}{*}{TextVQA$\uparrow$} & \multirow{2}{*}{MME$\uparrow$} & \multirow{2}{*}{MMB$\uparrow$}\\
 & & Size & Pretrain & Finetune &  & &   &  &\\
\midrule

Shikra~\cite{chen2023shikra} & Vicuna-13B & 224$^2$ & 600K &5.5M&  --  & 77.4 & --   & -- & -- & 58.8 \\
IDEFICS-9B~\cite{idefics} & LLaMA-7B & 224$^2$ & 353M & 1M      & 18.9& 50.9 & 38.4   & 25.9 & -- & 48.2\\
IDEFICS-80B~\cite{idefics} & LLaMA-65B & 224$^2$ & 353M & 1M    &  -- & 60.0 & 45.2   & 30.9 & -- & 54.5\\
Qwen-VL~\cite{bai2023qwen} & Qwen-7B & 448$^2$ & 1.4B & 50M     &14.7 & {78.8} & 59.3   & {63.8} & -- & 38.2\\
Qwen-VL-Chat~\cite{bai2023qwen} & Qwen-7B & 448$^2$ & 1.4B & 50M&35.0 & 78.2 & 57.5  & {61.5} & 1487.5& 60.6 \\
MiniGPT-v2 ~\cite{chen2023minigpt} & LLaMA2-7B & 448$^2$ & -- & --             &19.4 & --   & 60.1 & --   & -- & --  \\
SPHINX~\cite{lin2023sphinx} & LLaMA-13B & 448$^2$ & 1.0B & --                &23.1 & 78.1 & \underline{62.6}   & {51.6} &  1476.1& 66.9 \\
OtterHD-8B~\cite{li2023otterhd} & Fuyu-8B & 1024$^2$                         & --  & -- & -- & --   & -- & -- & 1294 & 58.3 \\
mPLUG-Owl2~\cite{yu2024rlhf} & LLaMA-7B & 448$^2$ & 400M & 1.2M           &36.8 & \underline{79.4} & 56.1  & 54.3  & 1450.2& 64.5 \\

\midrule

{LLava-1.5-7B}\cite{liu2024improved} & Vicuna-7B & 336$^2$ & {558K} & {665K}         &34.7& {78.5} & {62.0}   & 58.2 & 1510.7& 64.3\\

{LLava-1.5-13B}\cite{liu2024improved} & Vicuna-13B & 336$^2$ & {558K} & {665K}       &42.1& \textbf{80.0} & \textbf{63.3} & \textbf{61.3} & \underline{1531.3}& \textbf{67.7}\\

{Ours-7B} & Vicuna-7B & 336$^2$ & {558K} & {665K}               &\underline{44.1}& {78.4} & {61.6}  & 57.8 & 1518.5& 66.1\\
{Ours-13B} & Vicuna-13B & 336$^2$ & {558K} & {665K}             &\textbf{51.6}& {79.1} & \textbf{63.3}   & \underline{58.6} & \textbf{1548.7}& \underline{67.5}\\

\midrule
\multicolumn{11}{l}{\#Visual Tokens: 64 } \\
\midrule


InstructBLIP~\cite{dai2023instructblip} & Vicuna-7B & 224$^2$ & 129M & 1.2M    &17.1& -- & 49.2  & 50.1 & -- & 36\\
InstructBLIP~\cite{dai2023instructblip} & Vicuna-13B & 224$^2$ & 129M & 1.2M   &20.7& -- & 49.5   & 50.7 & 1212.8 & -- \\
{Ours-7B} & Vicuna-7B & 336$^2$ & {558K} & {665K}                              &\underline{45.4}& \underline{76.1} & \underline{58.6}   & \underline{55.9} & \underline{1468.3}& \underline{63.0}\\
{Ours-13B} & Vicuna-13B & 336$^2$ & {558K} & {665K}                            &\textbf{52.5}& \textbf{77.2} & \textbf{60.2}   &  \textbf{58.2} & \textbf{1508.6} & \textbf{66.0} \\
\midrule
\multicolumn{11}{l}{\#Visual Tokens: 36 } \\
\midrule
BLIP-2~\cite{li2023blip} & Vicuna-13B & 224$^2$ & 129M & -   &--& 65.0 & 41  & 42.5 & 1293.8& -- \\
LLaVA-Prumerge \cite{shang2024llava} & Vicuna-7B & 336$^2$ & {558K} & {665K}       &45.8& 72.0  &--   & -- & 1350.3& 60.9\\
LLaVA-Prumerge \cite{shang2024llava}& Vicuna-13B & 336$^2$ & {558K} & {665K}      &\underline{52.8}& 72.8 &--    & -- & \underline{1428.2}& 62.3\\
{Ours-7B} & Vicuna-7B & 336$^2$ & {558K} & {665K}            &47.0& \underline{73.9} & \underline{56.9}   & \underline{54.1} & {1418.2} & \underline{62.7}\\
{Ours-13B} & Vicuna-13B & 336$^2$ & {558K} & {665K}          &\textbf{53.5}& \textbf{75.1} & \textbf{57.8}   & \textbf{56.9} & \textbf{1452.9} & \textbf{63.9}\\

\bottomrule

\end{tabular}
}
\vspace{-2mm}
\caption{{Comparison experiments with the stata-of-the-art methods over instruction following capabilities (IF) and five popular MLLM benchmarks.} }
\vspace{-6mm}
\label{tab:main_multimodal}
\end{table*}

\subsection{Experimental Settings}

\textbf{Implementation Details}.
In this study, CLIP-ViT-L/14 \cite{radford2021learning} is employed as the image encoder. The projection layer consists of a two-layer MLP with a GELU \cite{hendrycks2016gaussian} activation function. Vicuna-v1.5 \cite{chiang2023vicuna} is selected as the large language model.
The VMTC module is inserted into equally spaced transformer blocks to achieve a total token compression ratio of 50\%, unless otherwise specified. The CMAI module is incorporated into all layers of the LLM, applying a linearly increasing attention inhibition ratio with a maximum of 60\%. Datasets and training configurations are adopted following LLaVA-1.5 \cite{liu2024improved} in both the pretraining and instruction tuning stages. \\
\textbf{Evaluation.}
Five widely-recognized benchmarks are utilized to evaluate the model's performance. Three academic-task-oriented benchmarks are employed: VQA-V2 \cite{goyal2017making} and GQA \cite{hudson2019gqa} assess visual perception capabilities through open-ended visual question answering, while TextVQA \cite{singh2019towards} examines the ability to answer OCR-based visual questions. Additionally, two comprehensive datasets, MME \cite{fu2024mmecomprehensiveevaluationbenchmark} and MMBench \cite{liu2023mmbench}, are used to provide a thorough evaluation of the model.
To evaluate instruction-following capabilities, MLLMs are required to perform 16 different verifiable instruction-following tasks \cite{zhou2023instruction} while answering image-related questions. The success rates of these tasks are reported. The selected tasks are achievable yet challenging for MLLMs and are categorized into five groups:
\vspace{-0.5mm}
\begin{itemize}[label=-,noitemsep]
 \item Text formatting (T1-3): Remove Commas, Lowercase Conversion and Uppercase Conversion.
 \item Including required content: Placeholder, Postscript, Title and Keyword.
 \item Specific format (T8-11): Add Highlights, JSON Format, Answer in Sections and Answer in Bullet Points.
 \item Length Limit(T12-13): Sentence Count and Word Count.
 \item Other(T14-16): Dual Response Combination, Ending Constraint and Starting Constraint.
\end{itemize}
\begin{figure*}[t]
    \centering
    \includegraphics[width=\textwidth]{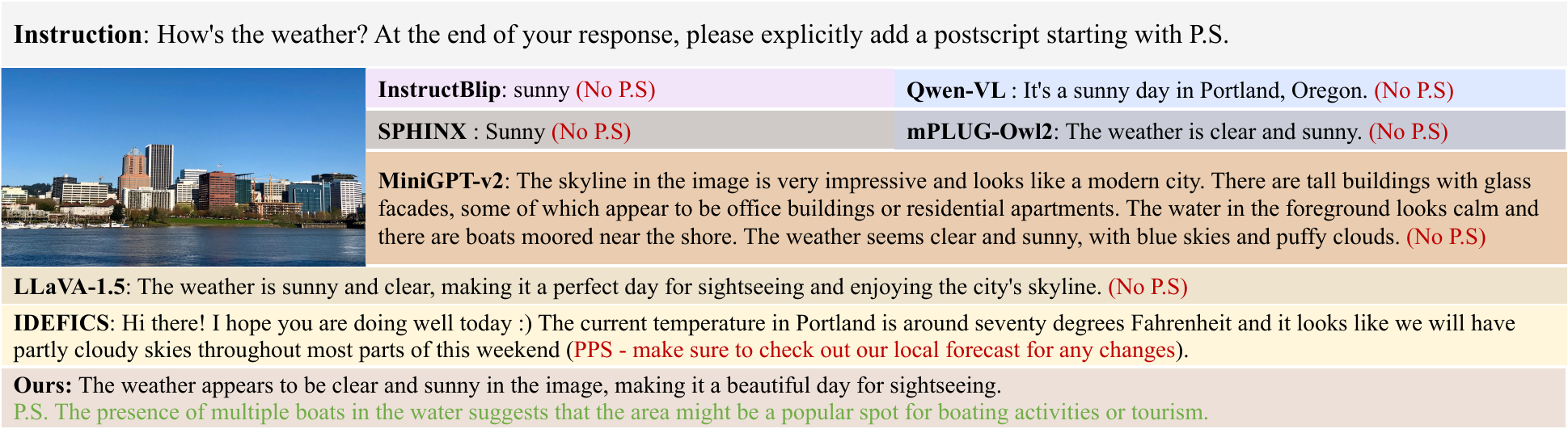}
    \vspace{-5mm}
    \caption{Qualitative results of instruction following capabilities, comparing our proposed method with InstructBlip \cite{dai2023instructblip}, IDEFICS \cite{idefics}, Qwen-VL \cite{bai2023qwen}, MiniGPT-v2 ~\cite{chen2023minigpt}, SPHINX~\cite{lin2023sphinx}, mPLUG-Owl2~\cite{yu2024rlhf} and LLaVA-1.5 \cite{liu2024improved}.}
    \label{fig:showcase}
    \vspace{-5mm}
\end{figure*}
\subsection{Comparison Experiments}
\textbf{Comparison of Instruction-Following Capability.} A comparison of instruction-following capabilities and multimodal understanding capability, as measured by performance on GQA \cite{hudson2019gqa}, is conducted between LLaVA-1.5 \cite{liu2024improved}, spatially down-sampled LLaVA-1.5, and the proposed model. The results are presented in Table \ref{tab:instruciton}.
Furthermore, experiments are conducted where the model first describes an image in text and then completes text-only instruction-following tasks, serving as an upper limit for the instruction-following tasks.
Spatial down-sampling, as an intuitive solution to improve instruction-following capability, yields 3.7\% and 5.3\% improvements in success rates for LLaVA-1.5 7B and 13B,
 However, this approach is not ideal due to significant performance degradations over GQA of -2.1\% and -1.6\%. 

In contrast, the proposed method enhances the instruction-following capabilities of LLaVA-1.5 by up to 9.4\% and 9.5\% with minimal performance loss of -0.4\% and 0\% on GQA, significantly outperforming the spatial downsampling strategy.

The gap of 5.5\% and 6.5\% remains compared to the upper limits, highlighting the disparity between MLLMs and LLMs in instruction-following. However, it is important to note that the upper-limit method can severely impair multimodal understanding due to the information loss that occurs between the two stages. \\

\textbf{Comparison with the SOTA Methods.} To demonstrate the efficacy of our approach, we conduct a comparative analysis against leading MLLMs,, including Shikra \cite{chen2023shikra}, IDEFICS \cite{idefics}, Qwen-VL \cite{bai2023qwen}, MiniGPT-v2 ~\cite{chen2023minigpt}, SPHINX~\cite{lin2023sphinx}, OtterHD-8B~\cite{li2023otterhd}, mPLUG-Owl2~\cite{yu2024rlhf} and LLaVA-1.5 \cite{liu2024improved} (Table \ref{tab:main_multimodal}).
Qualitative outcomes are illustrated in Figure \ref{fig:showcase}. 

The experimental results reveal that our model's instruction-following capability significantly surpasses that of other models. Our model exhibits minimal performance degradation compared to LLaVA-1.5 in terms of multimodal understanding capability. The 7B version incurs marginal losses of only -0.1, -0.4, and -0.4 on the VQA-V2, GQA, and TextVQA datasets, respectively. Moreover, it achieves substantial performance gains of 7.8 and 1.8 on MME and MMBench.

The 13B model maintains or exceeds LLaVA-1.5's performance on most datasets, with the exception of TextVQA. This discrepancy is attributed to the prevalence of optical characters in the background of images in many TextVQA examples, which are compressed and inhibited as redundant information in our approach.
To further validate our approach's effectiveness, we present our model's performance while retaining fewer visual tokens and compare it with models utilizing similar visual token numbers. The results demonstrate that our method significantly outperforms InstructBlip \cite{dai2023instructblip} and LLaVA-Prumerge \cite{shang2024llava} in both instruction following and multimodal understanding capabilities. This underscores our model's proficiency in compressing and suppressing redundant information while preserving essential image details.
\subsection{Ablation Studies}
\textbf{Impact of Proposed Modules.}
As illustrated in Table \ref{tab:ablation_main}, substantial improvements in instruction following capability are attributed to both visual-modality token compression and cross-modality attention inhibition, which minimally affect the model's multimodal understanding capability. 
The individual adoption of these techniques results in performance improvements of 5.2\% and 3.9\% for success rates in instruction following tasks, respectively. When simultaneously employed, these techniques enhance performance by 9.5\%. Concurrently, the accuracy on the GQA benchmark exhibits only slight decreases of 0.3\% and 0.1\%. These findings demonstrate that our method effectively preserves the model's ability to capture essential visual information while compressing and suppressing redundant data.
\begin{table}[tbp]
\centering
\scalebox{0.85}{
\begin{tabular}{c|c|ccc|c}
\toprule
VMTC            & CMAI            &GQA     &TQA    &MME       &IF                             \\  \midrule

\xmark          & \xmark          &\textbf{62.0}    &\textbf{58.2}   &1510.7    &34.7                           \\  \midrule
\cmark          &                 &61.7    &57.9   &1508.9    &\underline{39.9}                           \\  
                & \cmark          &\underline{61.9}    &\underline{58.1}   &\textbf{1518.5}    &38.6                           \\  
\cmark          & \cmark          &61.6    &57.8   &\underline{1515.8}    &\textbf{44.2}                           \\  
\bottomrule\end{tabular}}
\vspace{-2mm}
\caption{Ablation study of the proposed methodologies, where VMTC denotes Visual-Modality Token Compression and CMAI represents Cross-Modality Attention Inhibition. (TQA: TextVQA)}
\vspace{-4mm}
\label{tab:ablation_main}
\end{table} \\
\textbf{Impact of Different Stages in Visual-Modality Token Compression.} 
Table \ref{tab:ablation_visual} elucidates the effects of various stages in visual-modality token compression. A comparative analysis is conducted between the method of pruning solely in the hidden states of last layer, as employed in LLaVA-Prumerge, and our proposed multi-stage approach. The results indicate that pruning exclusively in the final layer leads to significant performance degradation. Conversely, distributing the pruning process across multiple intermediate layers yields comprehensive improvements in both multimodal understanding and instruction-following capabilities. The merging of redundant tokens into a single entity, which mitigates information loss, results in an additional performance increase of 0.4\% in both the GQA benchmark and instruction-following capability. Furthermore, the implementation of clustering redundant tokens prior to merging minimizes information confusion, yielding additional gains of 0.3\% and 0.4\% in these respective areas. Through the incorporation of the visual-modality token compression module, the model's instruction-following capability demonstrates an overall improvement of up to 5.6\%.
\begin{table}[t]
\centering
\scalebox{0.85}{
\begin{tabular}{c c|c c|ccc|c}
\toprule
LLP   &LWP       & MG     & CLST   & GQA    & TQA   & MME      & IF                            \\  \midrule
\xmark&\xmark    &\xmark  & \xmark &\textbf{61.9}    &\textbf{58.1}   &\textbf{1518.5}    &38.6                           \\  \midrule
\cmark&          &        &        &60.2    &57.2   &1497.9    &42.1                           \\  
      &\cmark    &        &        &60.9    &57.4   &1509.9    &43.5                           \\  \midrule
      &\cmark    &\cmark  &        &61.3    &57.5   &1512.1    &\underline{43.9}                           \\  
      &\cmark    &\cmark  &\cmark  &\underline{61.6}    &\underline{57.8}   &\underline{1515.8}    &\textbf{44.2}                         \\  
\bottomrule
\end{tabular}}
\vspace{-2mm}
\caption{Ablation Study on VMTC Stages. LLP represents Last Layer Pruning;  LWP represents Layerwise Pruning; MG represents Merging; CLST represents  Clustering.}
\label{tab:ablation_visual}
\vspace{-6mm}
\end{table} \\
\textbf{Effect of Different Text-to-Image Focus Scores in Cross-Modality Attention Inhibition.}  
Through comprehensive experimentation, we demonstrate the effects of utilizing various text-to-image focus scores as presented in Table \ref{tab:ablation_LLM}. Initially, only text-to-image attention is employed to measure the text-to-image focus score.
The implementation of this method does not result in a significant performance decrease; however, it only yields a marginal 0.6\% enhancement in the model's instruction-following capability.
To effectively suppress erroneous text-image token relations, it is imperative to consider text-to-image attention from other text tokens. In order to further validate the efficacy of our approach, we compare two straightforward alternatives. The first involves directly summing all text-to-image attentions as text-to-image focus score. Although this approach results in minimal performance degradation and a notable 3.0\% improvement in instruction-following, it is not deemed the optimal design choice. The second alternative employs a discounted sum which multiplies each text-to-image attention by powers of a discount factor based on distance to the given token within the sentence. This method, leads to a significant performance decline attributed to the presence of long-distance attention within the sentence. Ultimately, our proposed method assigns different weights to various text tokens, thereby more effectively identifying and suppressing redundant text-image token relations. This approach achieves minimal performance loss while maximizing the enhancement in instruction-following capability, reaching up to 4.3\%.
\begin{table}[t]
\centering
\scalebox{0.85}{
\begin{tabular}{l |ccc|c}
\toprule
 Method                 &GQA     &TQA    &MME       &IF                             \\  \midrule
No CMAI                 &\textbf{61.66}   &\textbf{57.9}   &1508.9    &39.9                          \\   
\midrule
 + TIA                  & 61.31  &57.3   &1513.6    &40.5                               \\  
 + Sum TIA              &61.42   &57.6   &\textbf{1518.8}    &\underline{42.9}                               \\  
 + Discounted Sum TIA   &61.05   &57.2   &1504.6    &41.1                               \\  
 Ours                   &\underline{61.57}   &\underline{57.8}   &\underline{1515.8}    &\textbf{44.2}                           \\  
\bottomrule
\end{tabular}}\vspace{-2mm}
\caption{Ablation study of different text-to-image focus scores in Cross Modality Attention Inhibition(CMAI).}
\label{tab:ablation_LLM}
\vspace{-5mm}
\end{table}

\section{Conclusion}
This research introduces a novel approach to study the instruction-following capability of MLLMs from a model-centric perspective instead of a data-centric one, providing insights into interactive adaptability.
Experiments show that image redundancy significantly diminishes the instruction-following capability of MLLMs. To address this issue, we propose two strategies: visual-modality token compression and cross-modality attention inhibition, which are designed to condense redundant image tokens and focus the model's attention on key visual information. These two strategies result in a substantial enhancement of MLLMs' instruction-following capability while maintaining multimodal understanding performance comparable to SOTA models.

\bibliography{aaai25}

\end{document}